\documentclass[5p]{elsarticle}

\usepackage{cmap}
\usepackage{tabularx}
\usepackage[version=3]{mhchem}
\usepackage[symbol*,hang]{footmisc} 
\usepackage{pdflscape} 
\usepackage{afterpage}
\usepackage{url}
\usepackage[hidelinks]{hyperref} 
\usepackage[skip=-3pt]{caption}
\usepackage{ragged2e}

\newcommand{\fig}[7]{
\begin{figure}[#7]
\begin{center}
\includegraphics[page=#2, width=#3\columnwidth, trim=#4, clip=true, keepaspectratio]{#1}
\end{center}
\caption{#5}
\label{#6} 
\end{figure}
} 

\newcommand{\bigfig}[6]{
\afterpage{
\begin{landscape}
\begin{figure}[p!]
\centering
\includegraphics[page=#2, width=#3\columnwidth, trim=#4, clip=true, keepaspectratio]{#1}
\caption{#5}
\label{#6} 
\end{figure}
\end{landscape}
}} 

\newcommand{\argh}[0]{\textsuperscript{\textregistered}}

\newcommand{\degrees}[0]{\ensuremath{^\circ}}
\newcommand{\tildey}[0]{{\raise.17ex\hbox{$\scriptstyle\sim$}}}

\begin{document}

\begin{frontmatter}

\title{An Extreme Learning Machine Approach to Predicting\\Near Chaotic HCCI Combustion Phasing in Real-Time\tnoteref{t1}}

\tnotetext[t1]{There are no fundamental changes in this updated version of the original October, 2013 pre-print paper~\cite{arXiv}.  This version includes algebraic simplifications, minor corrections, and improved body text.}

\author[uofm]{Adam Vaughan\corref{corresponding_av}}
\ead{vaughana@umich.edu}

\author[uofm]{Stanislav V. Bohac}
\ead{sbohac@umich.edu}

\address[uofm]{Dept. of Mechanical Engineering, University of Michigan, Ann Arbor, MI 48109, USA}

\cortext[corresponding_av]{Corresponding author}


\begin{abstract}
Fuel efficient Homogeneous Charge Compression Ignition (HCCI) engine combustion timing predictions must contend with non-linear chemistry, non-linear physics, period doubling bifurcation(s), turbulent mixing, model parameters that can drift day-to-day, and air-fuel mixture state information that cannot typically be resolved on a cycle-to-cycle basis, especially during transients.  In previous work, an abstract cycle-to-cycle mapping function coupled with $\epsilon$-Support Vector Regression was shown to predict experimentally observed cycle-to-cycle combustion timing over a wide range of engine conditions, despite some of the aforementioned difficulties.  The main limitation of the previous approach was that a partially acausual randomly sampled training dataset was used to train proof of concept offline predictions.  The objective of this paper is to address this limitation by proposing a new online adaptive Extreme Learning Machine (ELM) extension named Weighted Ring-ELM.  This extension enables fully causal combustion timing predictions at randomly chosen engine set points, and is shown to achieve results that are as good as or better than the previous offline method.  The broader objective of this approach is to enable a new class of real-time model predictive control strategies for high variability HCCI and, ultimately, to bring HCCI's low engine-out \ce{NO_x} and reduced \ce{CO2} emissions to production engines.
\end{abstract}
\end{frontmatter}

\section{Introduction}
Since the 1800s, gasoline engines have largely been operated by (1) controlling power output with a throttle that restricts airflow, (2) using a simple spark to control burn timing, and (3) operating close to fuel-air stoichiometry for reliable spark ignition and so catalysts can reduce \ce{NO_x}, \ce{HC}, and \ce{CO} emissions. The throttle hurts fuel efficiency with pumping losses (especially at low-load), and the stoichiometric mixtures used are thermodynamically less fuel efficient than mixtures diluted with air or exhaust gases.

With the broad availability of enabling technologies (e.g. variable valve timing), a relatively new type of combustion called Homogeneous Charge Compression Ignition (HCCI) has received increased research interest over the past decade. HCCI uses autoignition to burn lean (excess air) mixtures and can produce ultra-low \ce{NO_x} quantities that do not require expensive catalyst aftertreatment. Instead of a spark, combustion timing is controlled by the thermodynamic trajectory of the mixture and complex chemical kinetics. With both ultra-low \ce{NO_x} production and freedom from the stoichiometric shackles of spark ignition, HCCI achieves greater fuel efficiency through thermodynamically ideal lean mixtures and unthrottled operation.  This improved fuel economy, has real-world relevance to near-term sustainability, national oil independence, and greenhouse gas initiatives that seek to curb petroleum usage.

The primary challenge of HCCI autoignition is to ensure that the burn timing is synchronized against the motion of the piston. This is important for efficient extraction of mechanical work from the fuel-air mixture and to avoid unsafe, noisy combustion or unstable (near chaotic) combustion oscillations. This synchronization is so important that combustion researchers do not use normal units of time. Instead, they use the angle of the crank, which (1) describes the position of the piston, and (2) represents time because each crank angle takes a certain amount of time at a fixed crank rotation speed.  The thermodynamic trajectory of the air-fuel mixture is driven by the piston varying the cylinder volume as function of crank angle (Fig.~\ref{newSchematic}). These angles are measured relative to when the piston is at the top of the cylinder, or Top Dead Center (TDC). In a four-stroke engine, TDC occurs twice per cycle. In different regions, the piston may be compressing or expanding the mixture, or, if a valve is open, moving the mixture into or out of the intake or exhaust manifolds.

\fig{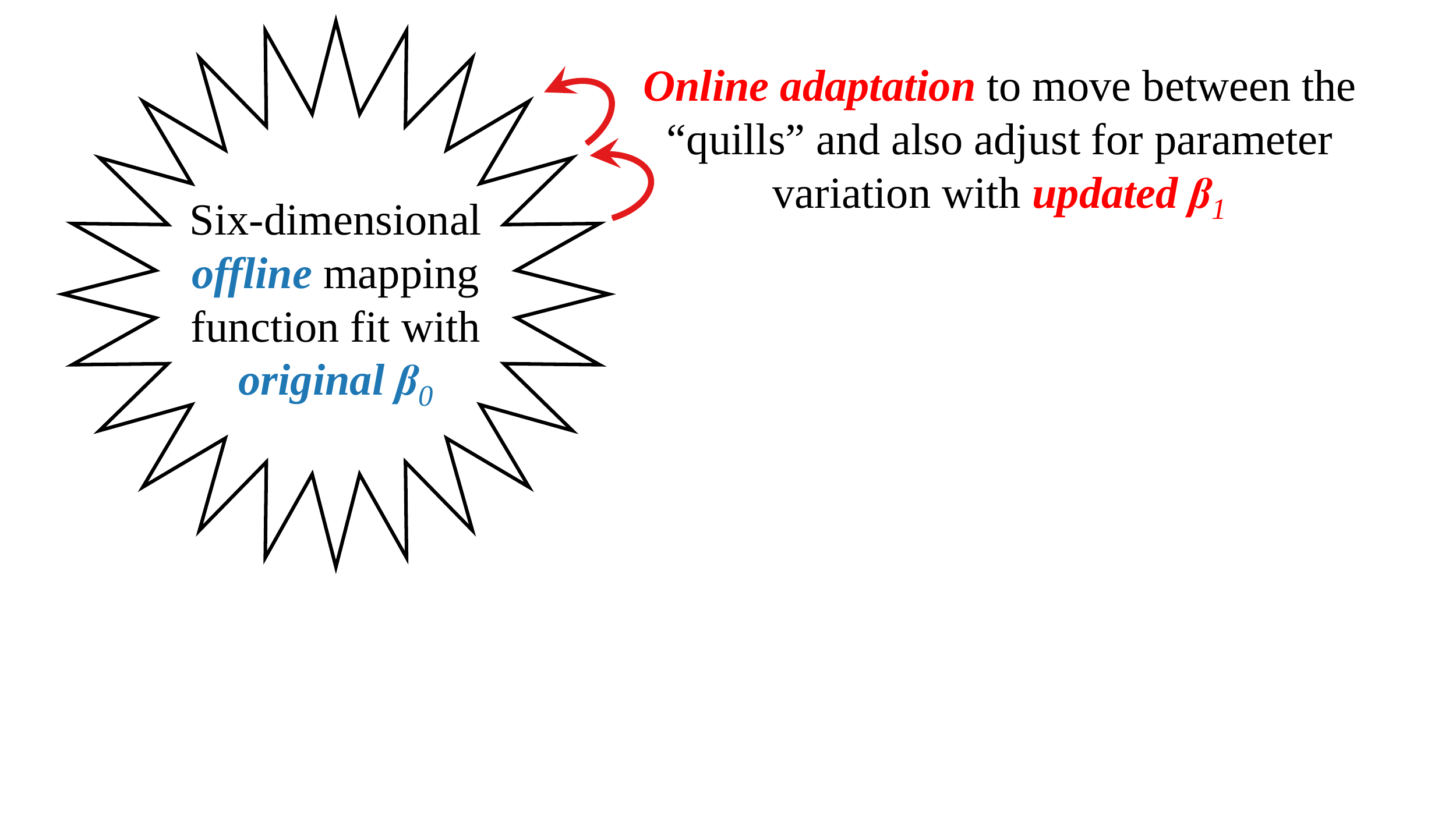}{2}{1.0}{0in 0in 2.85in 0in}{A schematic of key engine cycle variables.}{newSchematic}{t!} 

Highlighted on the cylinder volume curve are two regions, one for when the exhaust valve is open and the other for when the intake valve is open. Note that the two valve events are separated by a number of crank angle degrees, termed Negative Valve Overlap or NVO.  Unlike conventional engines, NVO prevents some of the hot exhaust gases from leaving the cylinder (typically 20-60\%~\cite{hcciCorrelationNvoRebreathingComparison,elliotResidualPaper}). This stores ``residual gases" for the next cycle, offering a practical way to raise the mixture temperature to ensure HCCI autoignition~\cite{lauraHcciIntakeHeatingPaper}. By changing the amount of NVO, one can affect the mixture temperature and dilution and ultimately control the chemical kinetics behind combustion timing.  Temperature and dilution work in opposite directions, but typically temperature dominates~\cite{saxenaReview}. NVO is not instantly adjustable with common variable valve timing systems, and the reader is cautioned that many researchers publish results with fully variable (lift and timing) electric or hydraulic valve actuation systems that are expensive to implement in production engines.

The use of NVO residual gases introduces strong cycle-to-cycle coupling on top of the already non-linear chemistry and physics that occur throughout a complete engine cycle~\cite{icefPaper}.  Further compounding the issues with residual gases is that neither the airflow to the cylinder(s) nor the quantity of residual gases in the cylinder can be accurately resolved before a burn happens on a {\it cycle-to-cycle} (not mean value) basis with commonly available sensors, especially during transients.  Beyond residual gas influences, there are also complex secondary influences on combustion behavior such as turbulent mixing, manifold resonance effects, combustion deposits, different varieties of fuel and even ambient temperature variations~\cite{joshFuelsAndDepositsThesis,vaughanContinuouslyVariableIntake}.

While HCCI is already a significant challenge given the above complexity, the combustion mode also exhibits a period doubling bifurcation cascade to chaos~\cite{icefPaper,dawSiHcci,erik}, similar to what is seen in high residual spark ignition engines~\cite{kantorDynamicInstabilitySi}.  When nearly chaotic, HCCI is still deterministic, but becomes oscillatory and very sensitive to parameter variations (e.g. residual gas fraction fluctuations~\cite{dawSiHcci,erik}).  This oscillatory ``stability limit" behavior is commonly referred to as high Cyclic Variability (CV) and it severely constrains the available load limits of HCCI.

\subsection{Motivation and goals}
A primary constraint for HCCI is the need to keep combustion timing between the ringing and combustion stability limits~\cite{lauraSaciPaper}.  At the ringing limit, excessive pressure rise rates are encountered, and at the stability limit, high CV in combustion timing is observed~\cite{lauraSaciPaper}.  Since these limits play a key role in constraining HCCI's usable operating range, it is desirable to explore new methods to predict the behavior at and beyond these constraints.  In particular, the ability to predict and correct for high CV might enable the use of late phased combustion to mitigate the excessive pressure rise rates that currently constrain HCCI's high-load operation~\cite{latePhasedCombustionIsGood}, while also potentially addressing the high CV experienced at low-load.  Towards the end goal of expanding the HCCI load envelope, this paper builds on previous work~\cite{icefPaper} by describing a new online adaptive machine learning method that enables fully causal cycle-to-cycle combustion timing predictions across randomly chosen engine set point transients that include both the stable and near chaotic bifurcation behavior.

\subsection{Experimental Observations}
In the authors' previous publication~\cite{icefPaper}, an abstract mapping function for engine combustion was created within the framework of a discrete dynamical system:
\begin{equation}
\begin{gathered}
\label{basicMappingFunction}
next \; combustion = \\
function( \;previous\;combustion,\; parameters)
\end{gathered}
\end{equation}
This simple abstraction is intended to convey a conceptual understanding of the experimental cycle-to-cycle behavior seen in Fig.~\ref{returnMaps}'s return maps.  These return maps show the experimentally observed combustion timing for a given cycle $n$ along the abscissa and the next cycle $n+1$ along the ordinate under random engine actuator set points~\cite{icefPaper}.  The value CA90 is the time in Crank Angle Degrees ({\degrees}CA) where 90\% of the fuel's net heat release is achieved, and thus measures the timing of the end of the burn in relation to piston's position as the crank rotates.\footnote{While it is not shown here, there is similar structure in the CA10 and CA50 percent burn metrics, although less pronounced (especially in CA10, see \cite{icefPaper}).}

\fig{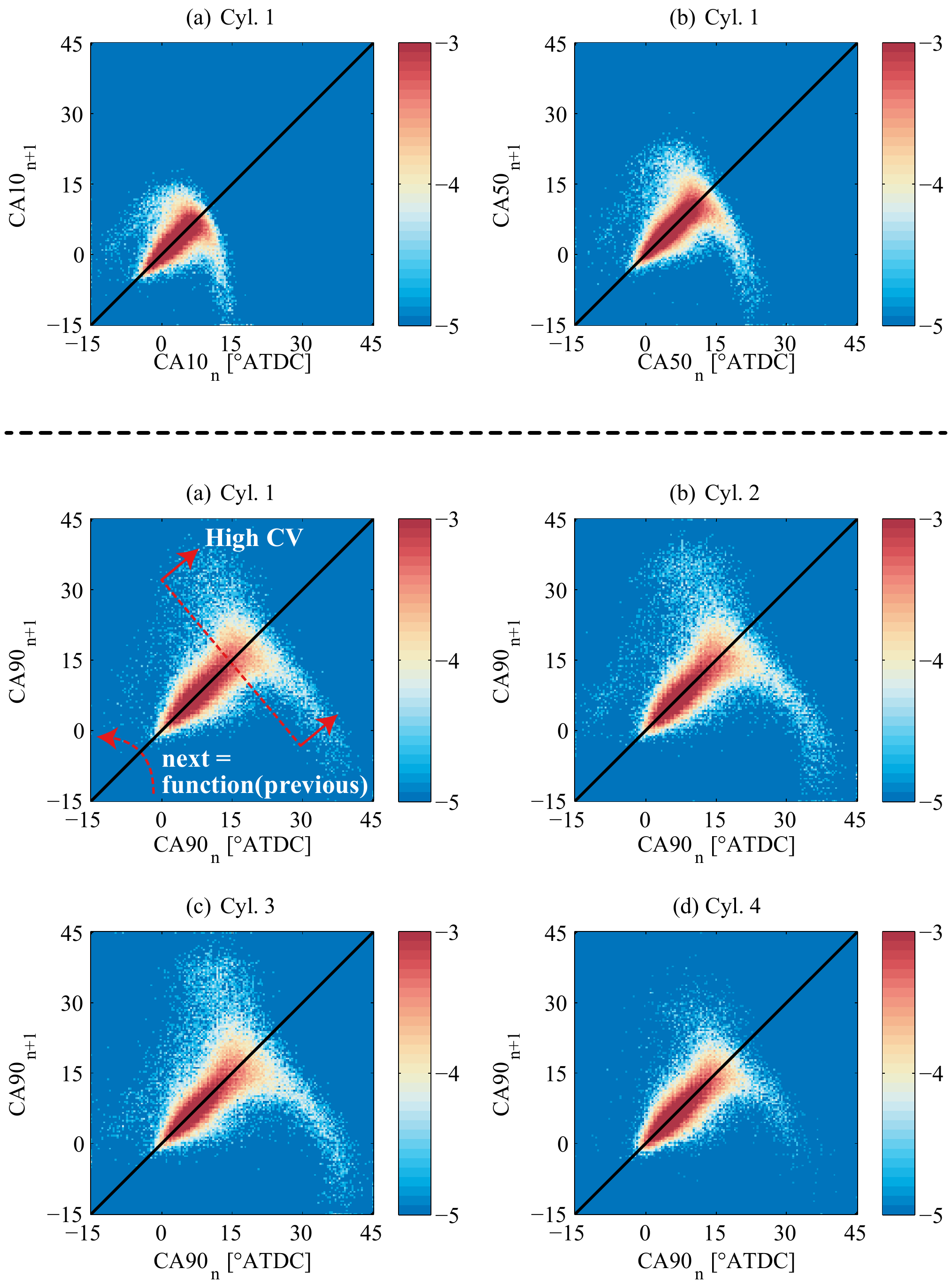}{1}{1.0}{0in 0in 0in 4.1in}{Return map probability histograms of CA90 generated from 129,964 cycles and 2,221 random engine set points.  Outliers are omitted and total only \tildey3\% of the data.  The colormap is $log_{10}$ to show order of magnitude differences.~\cite{icefPaper}}{returnMaps}{t!} 

The reader should note that there is structure to the cycle-to-cycle behavior despite the random actuator set points used to generate Fig.~\ref{returnMaps}.  This structure shows a deterministic transition to oscillatory high CV behavior as combustion moves towards later combustion timing that can be viewed as at least a single period doubling bifurcation with sensitive dependence on the engine set point~\cite{icefPaper,dawSiHcci,erik}. While mathematically interesting, this oscillatory high CV structure undesirably constrains practical HCCI engine operation.  A more thorough description of these data is provided in~\cite{icefPaper}.

\subsection{Modeling Approaches}
In~\cite{icefPaper}, a skeletal functional form for the abstract mapping function was built out using measurable quantities, thermodynamics, and known correlations.  Then, unlike the physics-based approaches that are usually discussed in the engine literature, the machine learning technique of $\epsilon$-Support Vector Regression ($\epsilon$-SVR) was combined with the abstract functional form to provide quantitative predictions.  The primary motivation for this machine learning approach was not that existing chemical kinetics with Computational Fluid Dynamics (CFD) cannot capture engine behavior (see~\cite{improvementOfPressureTracePredictions2009} and gasoline mechanism validation~\cite{ignitionDelaysOfGasoline}) but that the methods are:
\begin{itemize}
\item Too computationally intensive for real-time predictions.
\item Subject to experimental uncertainties in the {\it cycle-to-cycle} (not mean value) mixture state and composition.
\end{itemize}
As a point of reference, the simulation time of a 2,500 RPM, 48 millisecond engine cycle is measured in {\tildey}day(s) for a single core of a modern computer.

At the other computational complexity extreme, low-order approximation models of HCCI for control have been developed since at least the early 2000s~\cite{shaverThesis}, based on spark ignition engine knock models developed in the 1950s~\cite{livengoodAutoignitionPaper}.  Recently, efforts have been made to extend this type of model to the high CV regions of HCCI by injecting random residual gas fraction noise to capture uncertainties in the mixture state and composition~\cite{erik}.  This model was tuned for a limited set of steady-state conditions and only ``predictive'' in the sense that when injected with random (i.e. unpredictable) noise could it generate a qualitative return map shape similar to what is seen experimentally.  Time series predictions were not shown, only a cloud of possible combustion timings ranging from stable to oscillatory.  This fact was highlighted in an extension of the work~\cite{shyam} that again used random residual noise because ``the actual time series of disturbances to the experiments are unknown.''  That said, these models are useful for showing that a period doubling cascade to chaos driven by residual gas fraction can explain the observed high CV behavior.

In the context of the above, machine learning provides a computationally efficient way to capture complex combustion patterns while simultaneously avoiding explicit knowledge of the underlying mixture state and composition (provided an appropriate abstract mapping function is chosen).  While there are clearly benefits to this machine learning approach, a key issue is that machine learning is data driven, and relatively large quantities of data are needed to adequately cover large dimensional spaces. As shown conceptually in Fig.~\ref{porcupine}, these high dimensional data might be viewed as a ``porcupine''~\cite{learningFromData}.  Each engine operating condition might be viewed as a ``quill'' of Eq.~\ref{theModel}'s six-dimensional ``porcupine,'' and machine learning algorithms know nothing about the ideal gas law or chemical kinetics, so their ability to extrapolate between ``quills'' is limited, especially when provided sparse data.  Previous work~\cite{icefPaper} used a random sampling of cycle time series for training to ensure the data driven model had data to fit the ``quills,'' and then assessed the model's ability to predict on the remaining (randomly chosen) cycles.  Thus, the training dataset was partially acausual and that the model itself wasn't shown to adapt to new conditions.

\fig{newPaper_figures.pdf}{1}{1.0}{0.4in 1.7in 0.6in 0in}{High dimensional data might be viewed conceptually as a ``porcupine''~\cite{learningFromData}.  The primary goal of this paper is to design an online adaptive algorithm to fit new data between the ``quills.''}{porcupine}{h!} 

\subsection{Contribution}
The primary contribution of this work is the development of a new online learning method to provide real-time adaptive, fully causal predictions of near chaotic HCCI combustion combustion timing.  This method, called Weighted Ring - Extreme Learning Machine (WR-ELM), enables robust online updates to an Extreme Learning Machine (ELM) model that is trained to fit the ``quills'' of offline data.

\section{Methods}
\subsection{Mapping Function Modifications}
In previous work~\cite{icefPaper}, engine combustion was abstracted to the following mapping function:
\begin{equation}
\begin{gathered}
\label{theModel}
CA50_{n+1} = \\ function( \,CA90_n, \,TI, \,SOI, \,P_{IVC}, \,P_{EVO}, \,P_{NVO})
\end{gathered}
\end{equation}

\noindent where $n$ is the cycle iteration index, CA50 is the time in {\degrees}CA where 50\% of net heat release has occurred, CA90 is the time in {\degrees}CA when 90\% of net heat release has occured, $TI$ is the injection pulse width in milliseconds, $SOI$ is Start of Injection in {\degrees}CA Before Top Dead Center ({\degrees}BTDC), and the pressure variables measurements are mean pressures during specific regions of the combustion cycle (see Fig.~\ref{newSchematic}).{\setlength{\footnotemargin}{0pt}\footnote{Since CA90 is a stronger indicator of the oscillatory behavior seen in Fig.~\ref{returnMaps}, it is used as the model input.  CA50 is more commonly encountered in the engine literature and is thus used as the model output.  The two are related quantities, and in terms of model fit statistics there was no significant benefit of using one over the other.  That said, a few isolated instances were observed where CA90 did a better job predicting large oscillations.}}  Details of the simplified net heat release algorithm are available in~\cite{icefPaper}.  Fuel rail pressure is constant; however, the reader should note that the pressure drop to cylinder pressure during NVO injections varies with each transient step and during high CV regions.  The cylinder pressure variables $P_{IVC}$, $P_{EVO}$, and $P_{NVO}$ were chosen to capture cycle-to-cycle residual coupling and air flow without the difficulties of explicitly modeling those quantities.  To meet real-time engine controller timing requirements, $P_{IVC}$ and $P_{NVO}$ have been modified from~\cite{icefPaper}.  $P_{IVC}$ has been moved to the previous cycle, and the range of $P_{NVO}$'s mean has been shortened.  $P_{IVC}$ has also been moved closer to TDC to take advantage of the inherent signal amplification provided by the compression process.  The subscripts IVC, EVO, and NVO refer to the general timing regions Intake Valve Close, Exhaust Valve Open, and Negative Valve Overlap, respectively.

\subsection{WR-ELM Overview}
The primary benefits of an ELM approach over the $\epsilon$-SVR method used in~\cite{icefPaper} are:
\begin{itemize}
\item An ELM is easily adapted to online adaptation~\cite{osELM}.
\item An ELM provides good model generalization when the data are noisy~\cite{elmIntro}.
\item An ELM is extremely computationally efficient~\cite{osELM,elmIntro}.
\end{itemize}

\noindent WR-ELM is developed in this work as a weighted least squares extension to Online Sequential - ELM~\cite{osELM}.  While developed independently, a similar derivation is available in~\cite{wosELM}.  The difference between this work and the classification application in~\cite{wosELM} is the use of a ring buffer data structure ``chunk'' for online updates to an offline trained regression model.

The data in the WR-ELM ring buffer can be weighted more heavily than the data originally used to fit the offline model.  This allows emphasis to be placed on recent measurements that might be between the ``quills'' of Fig.~\ref{porcupine}'s offline trained model or the result of day-to-day engine parameter variation.  Thus, this approach allows one to prescribe a partitioned balance between the offline model fit against the need to adapt to the most recent conditions.  It also explicitly avoids over adaptation to the local conditions (that could compromise global generality) by ``forgetting'' old ring buffer data that eventually exit the buffer.  Fig.~\ref{ringBuffer} gives a schematic representation of this approach.

\fig{newPaper_figures.pdf}{3}{1.0}{0in 2.7in 0in 0in}{A schematic of overview WR-ELM.}{ringBuffer}{h!} 

Other differences from~\cite{osELM,wosELM} are that the derivation below lacks a bias vector $b$, uses the Gaussian distribution for $\textbf{a}$, drops the unnecessary logistic function exponential negative, and uses a Pad\'e approximant for $exp(x)$.  It was found empirically that the computation of the bias $b$ addition step could be removed with no loss of fitting performance if $\textbf{a}$'s elements were drawn from the Gaussian distribution $\mathcal{N}(0,1)$.  ELM theory only requires the distribution to be continuous~\cite{elmIntro}, although the ability to remove the bias is likely problem specific.

\subsection{WR-ELM Core Algorithm}

The basic goal of an Extreme Learning Machine (ELM) is to solve for the output layer weight vector $\boldsymbol\beta$ that scales the transformed input $\textbf{H}$ to output $\textbf{T}$:
\begin{equation}
\label{elmEquation}
\textbf{H}\boldsymbol\beta=\textbf{T}
\end{equation}
where $\textbf{H}$ is the hidden layer output matrix of a given input matrix and $\textbf{T}$ is the target vector.

For a set of $n$ input-output data pairs and $\widetilde{N}$ neurons at the $n$th cycle timestep, these variables are given by
\begin{equation}
\label{hiddenLayerOutputMatrix}
\textbf{H}(\textbf{a},\textbf{x}) =
\left[
\begin{matrix} 
G(\textbf{a}_1,\textbf{x}_1)&\hdots&G(\textbf{a}_{\widetilde{N}},\textbf{x}_1)\\
\vdots & \ddots & \vdots \\
G(\textbf{a}_1,\textbf{x}_n)&\hdots&G(\textbf{a}_{\widetilde{N}},\textbf{x}_n)\\
\end{matrix}
\right]_{n\times{\widetilde{N}}}
\end{equation}
where $G(\textbf{a}_i,\textbf{x})$ is the neuron activation function, chosen to be a commonly used logistic function, but without the unnecessary negative:
\begin{equation}
\label{logistic}
G(\textbf{a}_i,\textbf{x}) = \dfrac{1}{1+exp(\textbf{x} \cdot \textbf{a}_i )}~.
\end{equation}
Using a random input weight vector $\textbf{a}_i$ that is composed of random variable (r.v.) samples from a Gaussian distribution for each of the $z$ input variables gives
\begin{equation}
\label{}
\textbf{a}_i =
\left[
\begin{matrix} 
r.v.~\tildey~\mathcal{N}(0,1)
\\ \vdots
\\ r.v.~\tildey~\mathcal{N}(0,1)
\end{matrix}
\right]_{z \times 1}\;.
\end{equation}
The use of a random $\textbf{a}_i$ that is re-sampled for each of the $\widetilde{N}$ individual neurons during initialization is the main difference of an Extreme Learning Machine versus conventional neural networks that iteratively train each $\textbf{a}_i$~\cite{elmIntro}.  These $\textbf{a}_i$ vectors can then be collected into a single input weight matrix $\textbf{a}$, which is held fixed across all $n$ input row vectors $\textbf{x}$
\begin{equation}
\begin{aligned}
\label{inputMatrix}
&\textbf{x} =\\
&\left[
CA90_n\;\;TI\;\;SOI\;\;P_{IVC}\;\;P_{EVO}\;\;P_{NVO}
\right]_{n\times z}
\end{aligned}
\end{equation}
and $n$ output values
\begin{equation}
\label{outputVector}
\textbf{T} =
\left[
\begin{matrix} 
CA50_{n+1}
\end{matrix}
\right]_{n\times1}~.
\end{equation}

While the above logistic works well, one modification improves the computational efficiency on processors without a dedicated $exp(x)$ instruction, such as the Raspberry Pi\argh.  The modification is to replace the exponential with the following Pad\'{e} approximant:
\begin{equation}
exp(y) \approx p(y) = \dfrac{120 +60y +12y^2 +y^3}{120 -60y +12y^2 -y^3}\;,
\end{equation}
which has the following simple logistic relations:
\begin{equation}
\label{padeSigmoidA}
\dfrac{1}{1+exp(y)} \approx \dfrac{1}{1+p(y)} = \dfrac{(120 +12y^2) -60y -y^3}{2\cdot(120 +12y^2)}
\end{equation}
The small number of floating point operations used, reused intermediate terms, known boundedness of the normalized inputs, and known \textbf{a} weights make this approximant work well in this application.  No significant degradation in model performance was found, and as such it is used in all implementations described hereafter.

The normal equations can then be used to solve for the least squares solution ${\boldsymbol\beta}$ of Eq.~\ref{elmEquation} with
\begin{equation}
\label{elmNormalEquation}
{\boldsymbol\beta}=(\textbf{H}^\top\textbf{H})^{-1}\textbf{H}^\top\textbf{T}~.
\end{equation}
To extend this to a weighted least squares solution, one can incorporate a diagonal weight matrix $\textbf{W}$ to the normal equations~\cite{danSimon}:

\begin{equation}
\label{weightedNormalEquations}
{\boldsymbol\beta}=(\textbf{H}^\top\textbf{W}\textbf{H})^{-1}\textbf{H}^\top\textbf{W}\textbf{T}~.
\end{equation}

The solution can then be split between an offline and online ``chunk'' of recent input-output data pairs to avoid both the computational and storage burden of using offline data directly.  To do this, the matrices are partitioned with subscript 0 and 1 denoting the offline and online updated components, respectively:
\begin{equation}
\begin{gathered}
\label{}
\textbf{H}=
\left[
\begin{matrix} 
\textbf{H}_0
\\ \textbf{H}_1
\end{matrix}
\right]_{n\times{\widetilde{N}}}
\;
\textbf{W} =
\left[
\begin{matrix}
\textbf{W}_0&\textbf{0}
\\ \textbf{0}&\textbf{W}_1
\end{matrix}
\right]_{n\times N}
\\
\textbf{T}=
\left[
\begin{matrix} 
\textbf{T}_0
\\ \textbf{T}_1
\end{matrix}
\right]_{n\times 1}~.
\end{gathered}
\end{equation}
Then, following a similar derivation in~\cite{osELM} for recursive least squares but adding the weight matrix, the inversion portion $\textbf{H}^\top\textbf{W}\textbf{H}$ of the weighted normal equations Eq.~\ref{weightedNormalEquations} can be re-written in terms of $\textbf{K}_0$ and $\textbf{K}_1$:
\begin{equation}
\begin{split} 
\label{firstThird}
&\textbf{K}_1 = \\
&= \textbf{H}^\top\textbf{W}\textbf{H}
\\ &=
\left[
\begin{matrix} 
\textbf{H}_0
\\ \textbf{H}_1
\end{matrix}
\right]^\top
\left[
\begin{matrix} 
\textbf{W}_0&\textbf{0}
\\ \textbf{0}&\textbf{W}_1
\end{matrix}
\right]
\left[
\begin{matrix} 
\textbf{H}_0
\\ \textbf{H}_1
\end{matrix}
\right]
\\  &= 
\left[
\begin{matrix} 
\textbf{H}_0^\top &\textbf{H}_1^\top
\end{matrix}
\right]
\left[
\begin{matrix} 
\textbf{W}_0&\textbf{0}
\\ \textbf{0}&\textbf{W}_1
\end{matrix}
\right]
\left[
\begin{matrix} 
\textbf{H}_0
\\ \textbf{H}_1
\end{matrix}
\right]
\\  &= 
\left[
\begin{matrix} 
\textbf{H}_0^\top\textbf{W}_0 &\textbf{H}_1^\top\textbf{W}_1
\end{matrix}
\right]
\left[
\begin{matrix} 
\textbf{H}_0
\\ \textbf{H}_1
\end{matrix}
\right]
\\ &=\textbf{H}_0^\top\textbf{W}_0\textbf{H}_0 + \textbf{H}_1^\top\textbf{W}_1\textbf{H}_1
\\ &=\textbf{K}_0 + \textbf{H}_1^\top\textbf{W}_1\textbf{H}_1~.
\end{split}
\end{equation}
The non-inverted portion of the normal equations can similarly be re-written using existing relations:
\begin{equation}
\begin{split}
\label{secondThird}
& \textbf{H}^\top\textbf{W}\textbf{T} = \\
&= \left[
\begin{matrix} 
\textbf{H}_0
\\ \textbf{H}_1
\end{matrix}
\right]^\top
\left[
\begin{matrix} 
\textbf{W}_0&\textbf{0}
\\ \textbf{0}&\textbf{W}_1
\end{matrix}
\right]
\left[
\begin{matrix} 
\textbf{T}_0
\\ \textbf{T}_1
\end{matrix}
\right]
\\
&=\left[
\begin{matrix} 
\textbf{H}_0^\top & \textbf{H}_1^\top
\end{matrix}
\right]
\left[
\begin{matrix} 
\textbf{W}_0&\textbf{0}
\\ \textbf{0}&\textbf{W}_1
\end{matrix}
\right]
\left[
\begin{matrix} 
\textbf{T}_0
\\ \textbf{T}_1
\end{matrix}
\right]
\\
&=\left[
\begin{matrix} 
\textbf{H}_0^\top\textbf{W}_0 & \textbf{H}_1^\top\textbf{W}_1
\end{matrix}
\right]
\left[
\begin{matrix} 
\textbf{T}_0
\\ \textbf{T}_1
\end{matrix}
\right]
\\ & = \textbf{H}_0^\top\textbf{W}_0\textbf{T}_0 + \textbf{H}_1^\top\textbf{W}_1\textbf{T}_1
\\ & = \textbf{K}_0\textbf{K}_0^{-1}\textbf{H}_0^\top\textbf{W}_0\textbf{T}_0 + \textbf{H}_1^\top\textbf{W}_1\textbf{T}_1
\\ & = \textbf{K}_0{\boldsymbol\beta}_0 + \textbf{H}_1^\top\textbf{W}_1\textbf{T}_1 
\\ &= \left(\textbf{K}_1-\textbf{H}_1^\top\textbf{W}_1\textbf{H}_1\right) {\boldsymbol\beta}_0 + \textbf{H}_1^\top\textbf{W}_1\textbf{T}_1 
\\ & = \textbf{K}_1{\boldsymbol\beta}_0 -\textbf{H}_1^\top\textbf{W}_1\textbf{H}_1{\boldsymbol\beta}_0 + \textbf{H}_1^\top\textbf{W}_1\textbf{T}_1~.
\end{split}
\end{equation}

Substituting Eq.~\ref{secondThird} into the full online solution
\begin{equation}
\begin{split}
\label{rawOnline}
& {\boldsymbol\beta}_1 =
\\&= \left(\textbf{K}_1^{-1}\right)\left(\textbf{H}^\top\textbf{W}\textbf{T}\right)
\\& = {\boldsymbol\beta}_0  -\textbf{K}_1^{-1}\textbf{H}_1^\top\textbf{W}_1\textbf{H}_1{\boldsymbol\beta}_0 + \textbf{K}_1^{-1}\textbf{H}_1^\top\textbf{W}_1\textbf{T}_1 
\\& = {\boldsymbol\beta}_0  +\textbf{K}_1^{-1}\textbf{H}_1^\top\textbf{W}_1\left(\textbf{T}_1 -\textbf{H}_1{\boldsymbol\beta}_0 \right)
\end{split}
\end{equation}
yields the online solution without the need for the offline dataset.  To trade the computational burden of the $\widetilde{N}\times\widetilde{N}$ sized $\textbf{K}$ inverse for an inverse that scales with a smaller sized ring buffer, one can let $\textbf{P}=\textbf{K}^{-1}$
\begin{align}
\textbf{P}_0  &= \textbf{K}_0^{-1} = \left(\textbf{H}_0^\top\textbf{W}_0\textbf{H}_0\right)^{-1}
\\ \label{baseP1}
 \textbf{P}_1 &= \textbf{K}_1^{-1} = \left(\textbf{P}_0^{-1}+\textbf{H}_1^\top\textbf{W}_1\textbf{H}_1\right)^{-1}
\end{align}
and use the matrix inversion lemma on Eq.~\ref{baseP1} to yield:
\begin{equation}
\begin{gathered}
\label{inversionLemmaP1}
\textbf{P}_1 = \\
\textbf{P}_0-\textbf{P}_0\textbf{H}_1^\top\left(\textbf{W}_1^{-1}+\textbf{H}_1\textbf{P}_0\textbf{H}_1^\top    \right)^{-1}\textbf{H}_1\textbf{P}_0~.
\end{gathered}
\end{equation}

Unlike OS-ELM and WOS-ELM~\cite{osELM,wosELM}, additional simplification is possible because the WR-ELM algorithm does not propagate $\textbf{P}_1$ in time.  To begin, append the $\textbf{H}_1^\top\textbf{W}_1$ portion of Eq.~\ref{rawOnline} to Eq.~\ref{inversionLemmaP1} and distribute $\textbf{H}_1^\top$ to give:
\begin{equation}
\begin{gathered}
\label{substitutedP1}
\textbf{P}_1\textbf{H}_1^\top\textbf{W}_1 =
\\  \bigg[\textbf{P}_0\textbf{H}_1^\top-\textbf{P}_0\textbf{H}_1^\top\left(\textbf{W}_1^{-1}+\textbf{H}_1\textbf{P}_0\textbf{H}_1^\top    \right)^{-1}
\\ \cdot\textbf{H}_1\textbf{P}_0\textbf{H}_1^\top\bigg]\textbf{W}_1~.
\end{gathered}
\end{equation}
Eq.~\ref{substitutedP1} can then be simplified with the substitutions $\textbf{A} = \textbf{P}_0\textbf{H}_1^\top$, $\textbf{B} = \textbf{H}_1\textbf{A}$ and then distributing $\textbf{W}_1$ to provide:
\begin{equation}
\begin{gathered}
\label{abP1}
\textbf{P}_1\textbf{H}_1^\top\textbf{W}_1 =\\
\textbf{A}\left[\textbf{W}_1-\left(\textbf{W}_1^{-1}+\textbf{B}\right)^{-1}\textbf{B}\textbf{W}_1\right]~.
\end{gathered}
\end{equation}
Transforming Eq.~\ref{abP1} with the identity $(\textbf{X}+\textbf{Y})^{-1}\textbf{Y}$ $= \textbf{X}^{-1}(\textbf{X}^{-1}+\textbf{Y}^{-1})^{-1}$~\cite{matrixIdentities} gives:
\begin{equation}
\begin{gathered}
\label{abTransformedP1}
\textbf{P}_1\textbf{H}_1^\top\textbf{W}_1 = \\
\textbf{A}\left[\textbf{W}_1-\textbf{W}_1\left(\textbf{W}_1+\textbf{B}^{-1}\right)^{-1}\textbf{W}_1\right]~.
\end{gathered}
\end{equation}
Eq.~\ref{abTransformedP1} is then in a form where the identity $\textbf{X}-\textbf{X}(\textbf{X}+\textbf{Y})^{-1}\textbf{X}=(\textbf{X}^{-1}+\textbf{Y}^{-1})^{-1}$~\cite{matrixIdentities} can be applied to yield a substantially simpler form with a ring buffer sized inverse:
\begin{equation}
\label{abTransformedAgainP1}
\textbf{P}_1\textbf{H}_1^\top\textbf{W}_1 = \textbf{A}\left(\textbf{W}_1^{-1}+\textbf{B}\right)^{-1}~.
\end{equation}

Finally, noting that $\textbf{P}_1\textbf{H}_1^\top\textbf{W}_1 =  \textbf{K}_1^{-1}\textbf{H}_1^\top\textbf{W}_1$, one can then substitute Eq.~\ref{abTransformedAgainP1} into Eq.~\ref{rawOnline} and arrive at the following algorithm summary:

\begin{center}
\underline{OFFLINE TRAINING}
\end{center}
\begin{equation}
\begin{split}
\label{finalOffline}
\textbf{P}_0 &= \left[\left(\textbf{H}_0^\top\textbf{W}_0\textbf{H}_0\right)^{-1}\right]_{\widetilde{N}\times\widetilde{N}}
\\ {\boldsymbol\beta}_0 &=\left[\textbf{P}_0\textbf{H}_0^\top\textbf{W}_0\textbf{T}_0\right]_{\widetilde{N}\times 1}
\end{split}
\end{equation}

\begin{center}
\underline{ONLINE ADAPTATION}
\end{center}
\begin{equation}
\begin{gathered}
\label{finalOnline}
\textbf{A} = \textbf{P}_0\textbf{H}_1^\top,\; \textbf{B} = \textbf{H}_1\textbf{A}
\\ {\boldsymbol\beta}_1 = {\boldsymbol\beta}_0  +\textbf{A}\left(\textbf{W}_1^{-1}+\textbf{B}\right)^{-1} \left(\textbf{T}_1 -\textbf{H}_1{\boldsymbol\beta}_0 \right)
\end{gathered}
\end{equation}
\begin{center}
\underline{ONLINE PREDICTIONS}
\end{center}
\begin{equation}
\label{finalPredictions}
\text{CA50}_{n+2} = \textbf{T}_{n+1}= \textbf{H}(\textbf{a},\textbf{x}_{n+1}){\boldsymbol\beta}_1
\end{equation}
\noindent The reader should note that only $\textbf{P}_0$ and ${\boldsymbol\beta}_0$ are needed for online adaptation, and the size of these matrices scales only with an increasing number of neurons $\widetilde{N}$.  \textbf{None of the original offline data are needed}.  Additionally, note that Eq.~\ref{finalPredictions} is simply the reverse of Eq.~\ref{elmEquation} with the most recent $\textbf{x}_{n+1}$ cycle vector and ${\boldsymbol\beta}_1$ updated from the weighted ring buffer.  Finally, it should mentioned that the resulting update law Eq.~\ref{finalOnline} is structurally similar that of the steady-state Kalman filter~\cite{danSimon}, which also uses recursive least squares.  Future work should look at applying Kalman filtering algorithm improvements (e.g. square root filtering) to WR-ELM.

\subsection{Usage procedure}
\begin{enumerate}
\item Scale $\textbf{x}$ and $\textbf{T}$ columns between zero and unity for each variable. For the combustion implementation, column variable values below the 0.1\% and above the 99.9\% percentile were saturated at the respective percentile value, and then normalized between zero and unity between these percentile based saturation limits.  This was done to both adequately represent the distribution tails and avoid scaling issues.
\item The random non-linear transformation that enables the low computational complexity of the WR-ELM algorithm may result in ill-conditioned matrices.  All numerical implementations should use double precision.  Additionally, one should consider using Singular Value Decomposition for ill-conditioned matrix inversions.
\item Using the $\mathcal{N}(0,1)$ Gaussian distribution, initialize the $z \times \widetilde{N}$ ELM input weights $\textbf{a}$ and hold them fixed for all training / predictions.  For the combustion implementation, this was done with MATLAB\argh's~built-in $randn()$ function and the Mersenne Twister pseudo random number generator with seed 7,898,198.  An $\widetilde{N}$ of 64 was used based on initial trials, and each cylinder's individually computed WR-ELM model used an identical input weight matrix $\textbf{a}$.
\item Build $\textbf{H}_0(\textbf{a},\textbf{x}_{0})$  from previously acquired samples that cover a wide range of conditions with Eq.~\ref{hiddenLayerOutputMatrix} using an input matrix $\textbf{x}_0$ and output target vector $\textbf{T}_0$ (the formats of these are given in Eqs.~\ref{inputMatrix} and~\ref{outputVector}, respectively).  For the combustion implementation, the initial training data were \tildey40 minutes of random engine set points covering 53,884 cycles and 1,179 random engine set points at a single engine speed; however, it appears that only \tildey20 minutes of data may be sufficient.  Pruning the training data to only include \tildey6 cycles before and \tildey9 cycles after a transient set point step provided a small model fitting performance improvement.
\item \label{offlineWeightingItem} Specify a weight matrix $\textbf{W}_0$ for offline measurements.  For the combustion implementation, a simple scalar value $\textbf{W}_0= 3.5\times10^{-3}$ was chosen using a design of experiments.  While this weight works well as a proof of concept, future work should more rigorously determine the weight(s), perhaps with optimization techniques.  Note that $\textbf{W}_0$ allows weighting to be applied offline and that a small offline weighting is equivalent to a large online weighting.
\item Solve for the offline solution $\textbf{P}_0$ and ${\boldsymbol\beta}_0$ using Eqs.~\ref{finalOffline} and hold these values constant for all future predictions.
\item \label{onlineItem} Populate a ring buffer of size $r$ with recently completed input-output pairs using:
\begin{equation}
\begin{split}
\label{ringBufferStructure}
\begin{matrix}
\textbf{input} \\
\textbf{ring\;buffer}
\end{matrix}
 = \textbf{x}_{1} =& 
\left[
\begin{matrix} 
\textbf{x}_{n-r+1}
\\ \vdots
\\ \textbf{x}_{n}
\end{matrix}
\right]_{r\times z}
\\
\begin{matrix}
\textbf{output} \\
\textbf{ring\;buffer}
\end{matrix}
 = \textbf{T}_{1} =& 
\left[
\begin{matrix} 
CA50_{n-r+2}
\\ \vdots
\\ CA50_{n+1}
\end{matrix}
\right]_{r\times 1}~.
\end{split}
\end{equation}
\noindent Then execute the WR-ELM update algorithm between combustion cycle $n+1$ and $n+2$ as shown in Fig.~\ref{ringBuffer}.  For the combustion implementation, $r$ was taken to be 8 cycles after tuning with existing datasets.  If desired, $r$ can vary cycle-to-cycle.
\item As with the offline data, build $\textbf{H}_1(\textbf{a},\textbf{x}_{1})$ with Eq.~\ref{hiddenLayerOutputMatrix} using an input matrix $\textbf{x}_1$ and output target vector $\textbf{T}_1$.  Specify a weight matrix $\textbf{W}_1$.  For the combustion implementation the identity matrix ($\textbf{W}_1 = \textbf{I}$) was chosen since weighting was already applied to the offline data in step~\ref{offlineWeightingItem}.  Gradually increased weighting on the most recent time steps in the ring buffer was explored; however, it did not net a significant improvement to model fitting performance over a simple scalar value on offline data.  Although not explored in the current implementation, $\textbf{W}_1$ can vary cycle-to-cycle.

\item Solve for the updated ${\boldsymbol\beta}_1$ solution using Eqs.~\ref{finalOnline}.
\item \label{endStep} After cycle $n+1$'s input vector $\textbf{x}_{n+1}$ is fully populated, transform vector into $\textbf{H}_{n+1}$ using Eq.~\ref{hiddenLayerOutputMatrix} and solve for a predicted target value $\textbf{T}_{n+1}$ or CA50$_{n+2}$ using Eq.~\ref{finalPredictions}.
\item Repeat steps \ref{onlineItem}-\ref{endStep} for each new time step, caching results (e.g. hidden layer outputs) from previous time steps to reduce computational requirements.
\end{enumerate}

\subsection{Real-Time Implementation}
A collection of unoptimized MATLAB\argh~software routines was developed using the techniques described in the previous sections.  The offline solution provided by Eqs.~\ref{finalOffline} was solved at an average rate of 1.1 ${\mu}s$ per combustion cycle per cylinder on an Intel\argh~i7~860 2.8~GHz desktop computer running Gentoo GNU/Linux\argh.  The online predictions from Eqs.~\ref{finalOnline}, \ref{ringBufferStructure}, and~\ref{finalPredictions} were recast into a $parfor$ loop that automatically parallelized the code across four worker threads to provide predictions at an average rate of 66 ${\mu}s$ per combustion cycle per cylinder.  This level of performance is more than adequate for real-time.

Although algorithm development is the main focus of this paper, a real-time implementation of the WR-ELM algorithm has been built using custom, 18-bit Raspberry Pi\argh~data acquisition hardware.  A two-minute video demonstrating both predictions and control is available at~\cite{videoOfMyAlgorithm}.  The software for this system is comprised of:
\begin{itemize}
\item A PREEMPT\_RT patched Linux\argh~kernel with minor patches to completely disable Fast Interrupt reQuest (FIQ) usage by the USB driver.
\item ARM assembly code for high-speed pressure data acquisition using the FIQ (up to \tildey240 kilosamples per second, total all cylinder channels).
\item A C code Linux\argh~kernel module that contains the FIQ assembly code with page-based memory allocation and standard mmap, ioctl, and fasync hooks to Linux\argh~user space.
\item A multi-threaded Linux\argh~user space application that runs heat release calculations and WR-ELM.  This software leverages the Eigen C++ matrix library and a custom assembly code matrix multiply\footnote{This code was benchmarked to be 29\% faster than OpenBLAS's VFPv2 GEneric Matrix Multiply (GEMM) and 126\% faster than Eigen C++.} for the Raspberry Pi 1's VFPv2 double precision floating point unit.  Asynchronous fasync notification is used for specific crank angle events published from the FIQ code to synchronize the user space software's execution against the crank's rotation.
\item A low-priority second user space thread that uses standard WebSockets from the libwebsockets C code library to stream processed data to a web-based user interface coded in JavaScript with the d3.js library.
\item A minimal Raspbian GNU/Linux\argh~distribution~\cite{raspbian}.
\end{itemize}
After the adaptation routine is run, it is possible to perform 11 model predictive control predictions within a worst case task context switch and calculation latency window of \tildey300 ${\mu}s$.  This level of real-time performance is needed to ensure control authority with the $SOI$ actuator immediately after $P_{NVO}$ is measured.

\begin{table}[!b]
\caption{Experimental setup and test conditions}
\label{engineSpecs}
\begin{center}
\noindent
\small
\begin{tabularx}{\columnwidth}{X}
\hline \\[-0.75em]
Engine \\[-1pt]
\footnotesize \hspace{0.0125\columnwidth} Make / model \hfill GM / LNF Ecotec \\[-1pt]
\footnotesize \hspace{0.0125\columnwidth} Cylinder layout \hfill in-line 4 \\[-1pt]
\footnotesize \hspace{0.0125\columnwidth} Overall displacement \hfill 2.0 L \\[-1pt]
\footnotesize \hspace{0.0125\columnwidth} Bore / stroke \hfill 86 / 86 mm \\[-1pt]
\footnotesize \hspace{0.0125\columnwidth} Geometric compression ratio\textsuperscript{a}  \hfill 11.2 : 1 \\[-1pt]
\footnotesize \hspace{0.0125\columnwidth} Cam lift\textsuperscript{a}  \hfill 3.5 mm \\[-1pt]
\footnotesize \hspace{0.0125\columnwidth} Cam duration\textsuperscript{a}  \hfill 93 \degrees{CA} \\[-1pt]
\footnotesize \hspace{0.0125\columnwidth} Cam phaser type \hfill hydraulic \\[-1pt]
\footnotesize \hspace{0.0125\columnwidth} Fuel injector type \hfill direct, side mounted, wall guided \\

Fuel \\[-1pt]
\footnotesize \hspace{0.0125\columnwidth} Designation \hfill Haltermann HF0437, EPA Tier II EEE\\[-1pt]
\footnotesize \hspace{0.0125\columnwidth} Description \hfill U.S. Federal Emission Cert. Gasoline\\[-1pt]
\footnotesize \hspace{0.0125\columnwidth} Research Octane Number \hfill  97.0\\[-1pt]
\footnotesize \hspace{0.0125\columnwidth} Motor Octane Number \hfill  88.1\\[-1pt]
\footnotesize \hspace{0.0125\columnwidth} ASTM D240 heating value \hfill  42.8 MJ / kg \\[-1pt]
\footnotesize \hspace{0.0125\columnwidth} Aromatic / olefin / saturate fractions \hfill 28 / 1 / 71 \% volume\\

Test conditions \\[-1pt]
\footnotesize \hspace{0.0125\columnwidth} Throttle position \hfill wide open \\[-1pt]
\footnotesize \hspace{0.0125\columnwidth} Turbocharger \hfill wastegate open \\[-1pt]
\footnotesize \hspace{0.0125\columnwidth} Supercharger \hfill bypassed \\[-1pt]
\footnotesize \hspace{0.0125\columnwidth} Residual retention strategy \hfill negative valve overlap \\

\footnotesize \hspace{0.0125\columnwidth} IVO set point range\textsuperscript{b} \hfill 78.6 / 128 {\degrees}ATDC \\[-1pt]
\footnotesize \hspace{0.0125\columnwidth} EVC set point range\textsuperscript{b} \hfill -118 / -83.0 {\degrees}ATDC \\[-1pt]
\footnotesize \hspace{0.0125\columnwidth} $SOI$ set point range\textsuperscript{b} \hfill 272 / 378 {\degrees}BTDC \\[-1pt]
\footnotesize \hspace{0.0125\columnwidth} $TI$ set point range\textsuperscript{b} \hfill 0.582 / 1.01 ms \\

\footnotesize \hspace{0.0125\columnwidth} Net IMEP values visited\textsuperscript{b} \hfill 1.85 / 3.62 bar \\[-1pt]
\footnotesize \hspace{0.0125\columnwidth} Air-fuel ratios visited\textsuperscript{b} \hfill 0.90 / 1.6 \\[-1pt]
\footnotesize \hspace{0.0125\columnwidth} Estimated fuel per cycle\textsuperscript{b} \hfill 6 / 11 mg  \\

\footnotesize \hspace{0.0125\columnwidth} Intake runner temps., all cyls. \hfill $\mu = 52.6$ \degrees{C}, $\sigma = 1.6$ \degrees{C}\\[-1pt]
\footnotesize \hspace{0.0125\columnwidth} Fuel injection pressure \hfill $\mu = 70.0$ bar, $\sigma = 0.85$ bar\\[-1pt]
\footnotesize \hspace{0.0125\columnwidth} Coolant temperature \hfill $\mu = 89.5$ \degrees{C}, $\sigma = 3.4$ \degrees{C}\\[-1pt]
\footnotesize \hspace{0.0125\columnwidth} Engine speed \hfill $\mu = 2,500$ RPM, $\sigma = 6$ RPM\\[0.25em]

\hline\\[-4pt]
\textsuperscript{a} {\it Modified from stock engine} \\[1pt]
\textsuperscript{b} {\it First to 99\textsuperscript{th} percentile}
\end{tabularx}
\end{center}
\end{table}

\subsection{Experimental Setup}
Table~\ref{engineSpecs} provides a summary of the experimental setup and conditions visited.  In-cylinder pressure was acquired on a 1.0 {\degrees}CA basis and pegged thermodynamically for each cycle after IVC using a polytropic exponent of $1.35$.  This exponent was chosen to most closely match the pegging results achieved using the single intake runner high speed pressure sensor on cylinder~1.  For the purpose of computing cycle-to-cycle net Indicated Mean Effective Pressure (IMEP), a cycle was defined as starting at 360 {\degrees}BTDC firing and ending at 359 {\degrees}ATDC firing.  The reference cam lift for timing and duration in Table~\ref{engineSpecs} is 0.5 mm.  The air-fuel ratio range indicated in Table~\ref{engineSpecs} was measured post-turbine, and represents a mixture from all four cylinders.  Fuel mass per cycle was {\it estimated} using the fuel's lower heating value, assuming that the gross heat release was 20\% greater than the net heat release, and that the combustion efficiency was 100\%.

\subsection{Dataset Description}
The full collection of 129,964 cycles is comprised of five \tildey20 minute random test subsequences.  Each random subsequence covers the same nominal ranges listed in Table~\ref{engineSpecs}; however, one subsequence holds $SOI$ fixed.  The sequence with $SOI$ fixed is only used as part of Fig.~\ref{returnMaps}, and not during the model training and testing presented here.  Total cycle counts are reported after outliers are removed.  The outlier criteria (detailed in~\cite{icefPaper}) are intended to remove misfires and partial burns.  These criteria are fairly permissive and remove only \tildey3\% of the data.

The offline solution was trained using \tildey40 minutes of test cell time covering 53,884 cycles and 1,179 random engine set points at 2,500 rpm (two random subsequences).  These subsequences are comprised of random, transient set point steps occurring approximately every 0.5 - 10 seconds with occasional misfires covering the nominal variable ranges given in Table~\ref{engineSpecs}.  The training data were pruned to only include 6 cycles before and 9 cycles after a transient set point step for a small model fitting performance improvement.  The online solution was run with a separate random subsequence and fed unseen cycles one-by-one, similar to what would be experienced in a real-time implementation.   This online dataset is comprised of 25,323 consecutive cycles with 521 random engine set points.  Longer online sequences were also tested, and achieved similar results.

\section{Results and Discussion}
The fitting performance on a 25,323 cycle dataset (excluding outliers as defined in~\cite{icefPaper}) is shown in Table~\ref{goodnessOfFit} and in Figs.~\ref{histogram}, \ref{scatter}, and~\ref{betas}.  The minimum coefficient of determination ($R^2$) given in Table~\ref{goodnessOfFit} shows that at least 80\% of the cycle-to-cycle variance can be explained by the model as it is currently defined for a dataset with random transient steps occurring approximately every 0.5 - 10 seconds and occasional misfires.  This is better than the 76\% achieved with $\epsilon$-Support Vector Regression ($\epsilon$-SVR) on the same dataset in~\cite{icefPaper}.  However, this is not a 1:1 comparison because WR-ELM is fully predicting the entire 25,323 cycle dataset, whereas $\epsilon$-SVR's training strategy ensured the data-driven model had partially seen the operating points it was trying to predict.  Steady-state Root Mean Squared Error (RMSE) in Table~\ref{goodnessOfFit} was assessed at a single set point with a mean CA50 of 3.9 {\degrees}ATDC and a net Indicated Mean Effective Pressure (IMEP) of 2.8 bar before the transient sequence started.

\begin{table}[!b]
\small
\caption{WR-ELM model of Eq.~\ref{theModel} error statistics}
\begin{center}
\label{goodnessOfFit}
\begin{tabular}{c c c c}
\hline \\[-1em]
\centering
Cylinder & Overall$^c$  & Overall             & Steady-State \\
\#         &  $R^2$ & RMSE [{\degrees}CA] & RMSE [{\degrees}CA]  \\
\hline
1 & 0.81 & 1.85 & 0.84 \\
2 & 0.81 & 2.06 & 0.97 \\
3 & 0.80 & 2.17 & 0.97 \\
4 & 0.83 & 1.64 & 0.86 \\
\hline
\end{tabular}\\[8pt]
\textsuperscript{c} 25,323 consecutive cycles with random transient steps\\ occurring approx. every 0.5 - 10 sec. and occasional misfires.
\end{center}
\end{table}

\fig{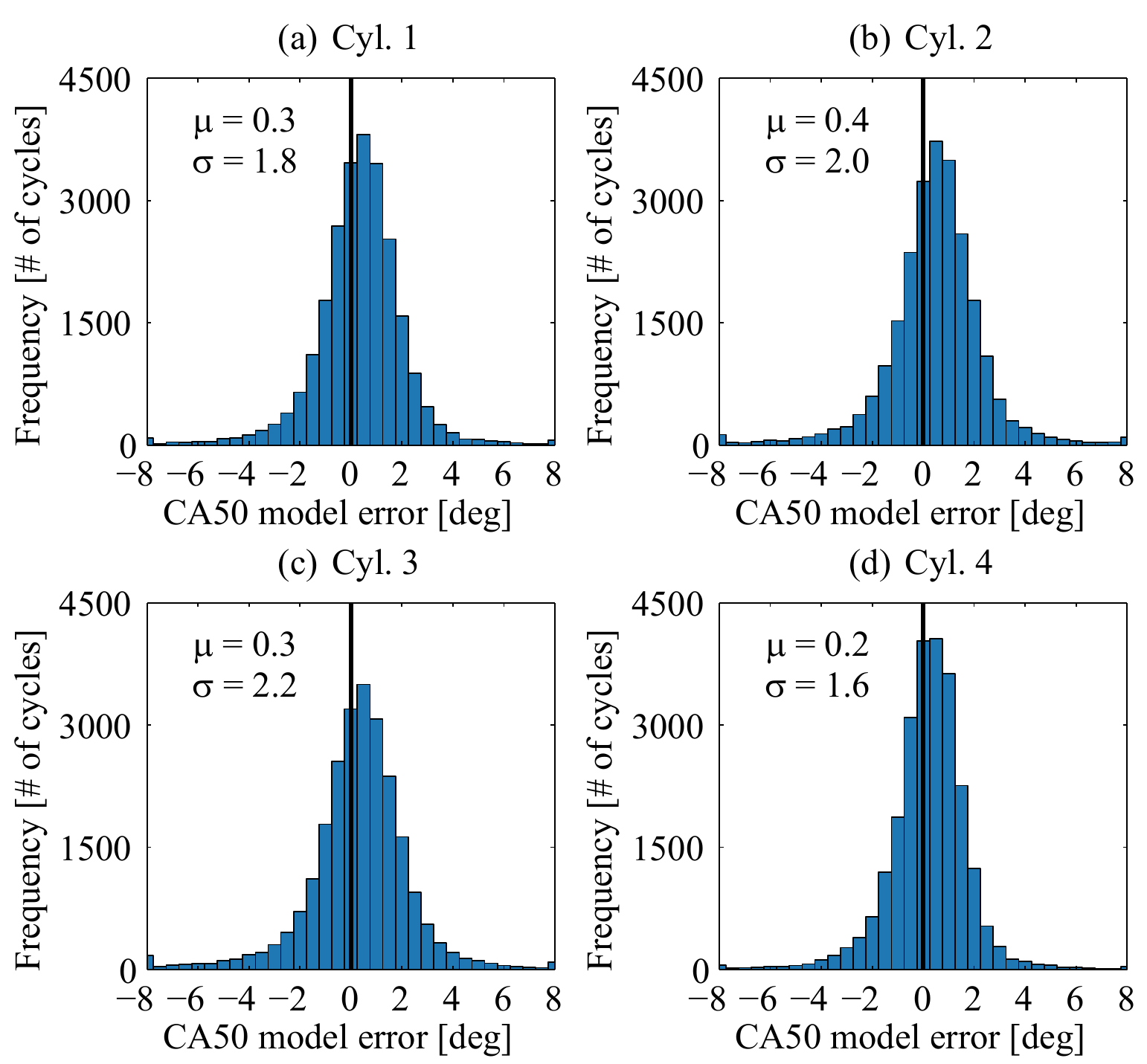}{1}{0.95}{0in 0in 0in 0in}{Error histograms for WR-ELM model of Eq.~\ref{theModel} across 25,323 consecutive cycles with random transient steps occurring approx. every 0.5 - 10 seconds and occasional misfires.}{histogram}{!h} 

\fig{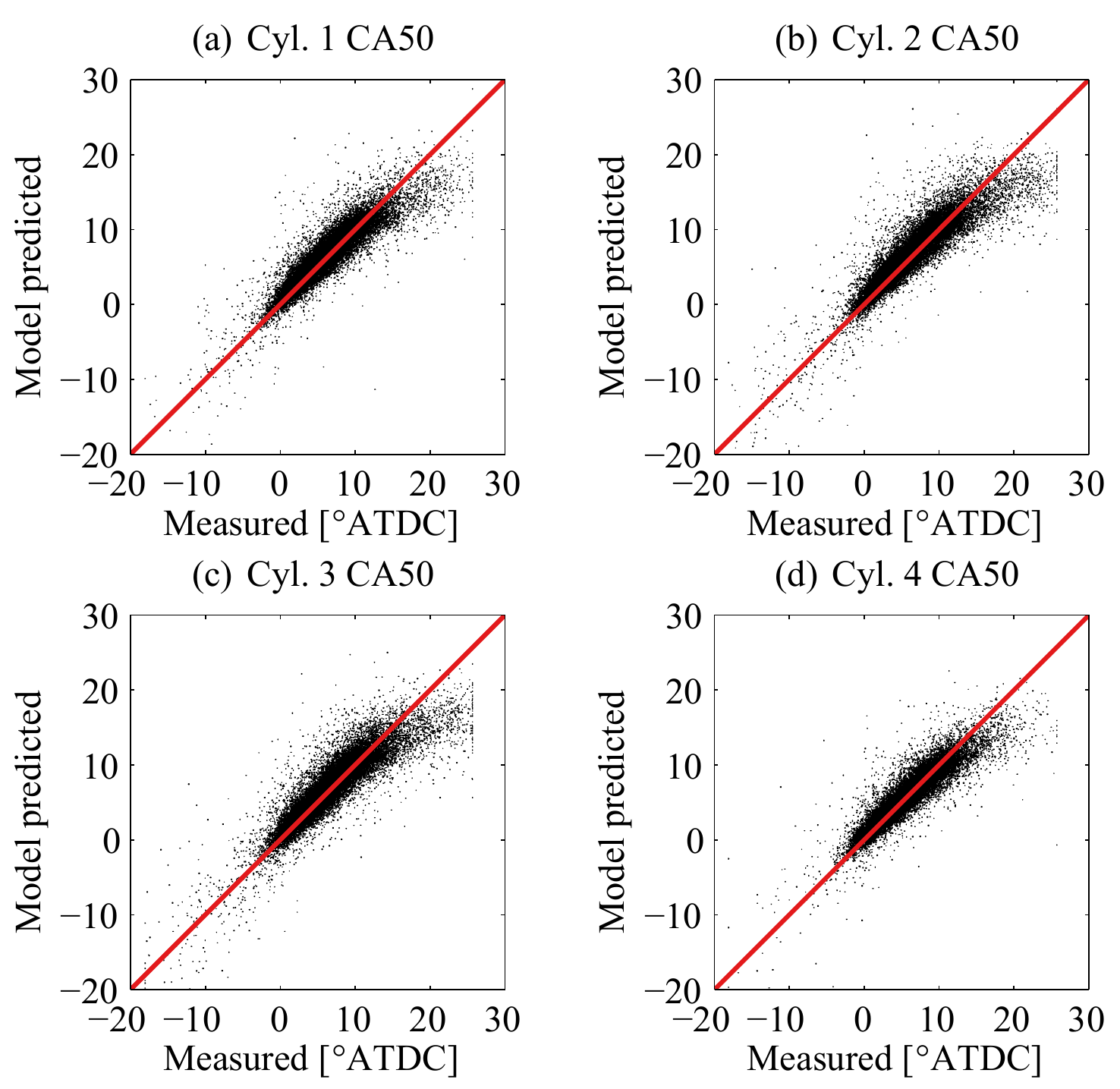}{1}{0.95}{0in 0in 0in 0in}{Predicted versus measured WR-ELM model of Eq.~\ref{theModel} across 25,323 consecutive cycles with random transient steps occurring approximately every 0.5 - 10 seconds and occasional misfires.  Late combustion timing is under predicted, but almost all prediction outliers capture the correct directionality.}{scatter}{!h} 

Fig.~\ref{histogram} shows the distribution of model errors.  It is clear that there is a slight positive bias to the predictions.  Fig.~\ref{scatter} provides insight into the tails of Fig.~\ref{histogram} and shows that model errors still generally capture the correct directionality.  Fig.~\ref{scatter} also shows that late combustion timing is under predicted and that the positive bias is largely from the midrange values of CA50.  Fig.~\ref{betas}a-d shows the cycle-to-cycle time series predictions, which can be computed as early as 358 {\degrees}BTDC firing.  Missing segments are the outliers described earlier.  Fig.~\ref{betas}e-h provide qualitative insight into the model's ${\boldsymbol\beta}_1$ weights under online adaptation.  The neurons are sorted by the 2-norm of their respective input weight vector $\textbf{a}_i$.  The same non-linear transformation specified by $\textbf{a}$ is used for each cylinder, and any cylinder-to-cylinder differences in the cycle-to-cycle ${\boldsymbol\beta}_1$ are due to different characteristics of each cylinder.  Fig.~\ref{betas}i shows the IMEP and Fig.~\ref{betas}j-l shows the random engine actuator inputs that are driving each engine set point transient.

The model predictions of Fig.~\ref{betas}a-d generally show good agreement; however, there are occasional tracking errors.  It is unclear what the source of these tracking errors is (e.g. is it a fundamental model limitation, the need for more inputs,\footnote{Adding additional inputs is not necessarily practical computationally, experimentally, or even advisable given Occam's razor.} the influence of other misfiring cylinders going through a harsh reignite, the need for more weight tuning or offline training data, or perhaps something else?).  Future work will try to answer these questions.  Overall, however, the authors believe the level of fit shown in Table~\ref{goodnessOfFit} and in Figs.~\ref{histogram}, \ref{scatter}, and~\ref{betas} is very good considering that the dataset includes both transients and operating points with high CV, right up to complete engine misfire.

\bigfig{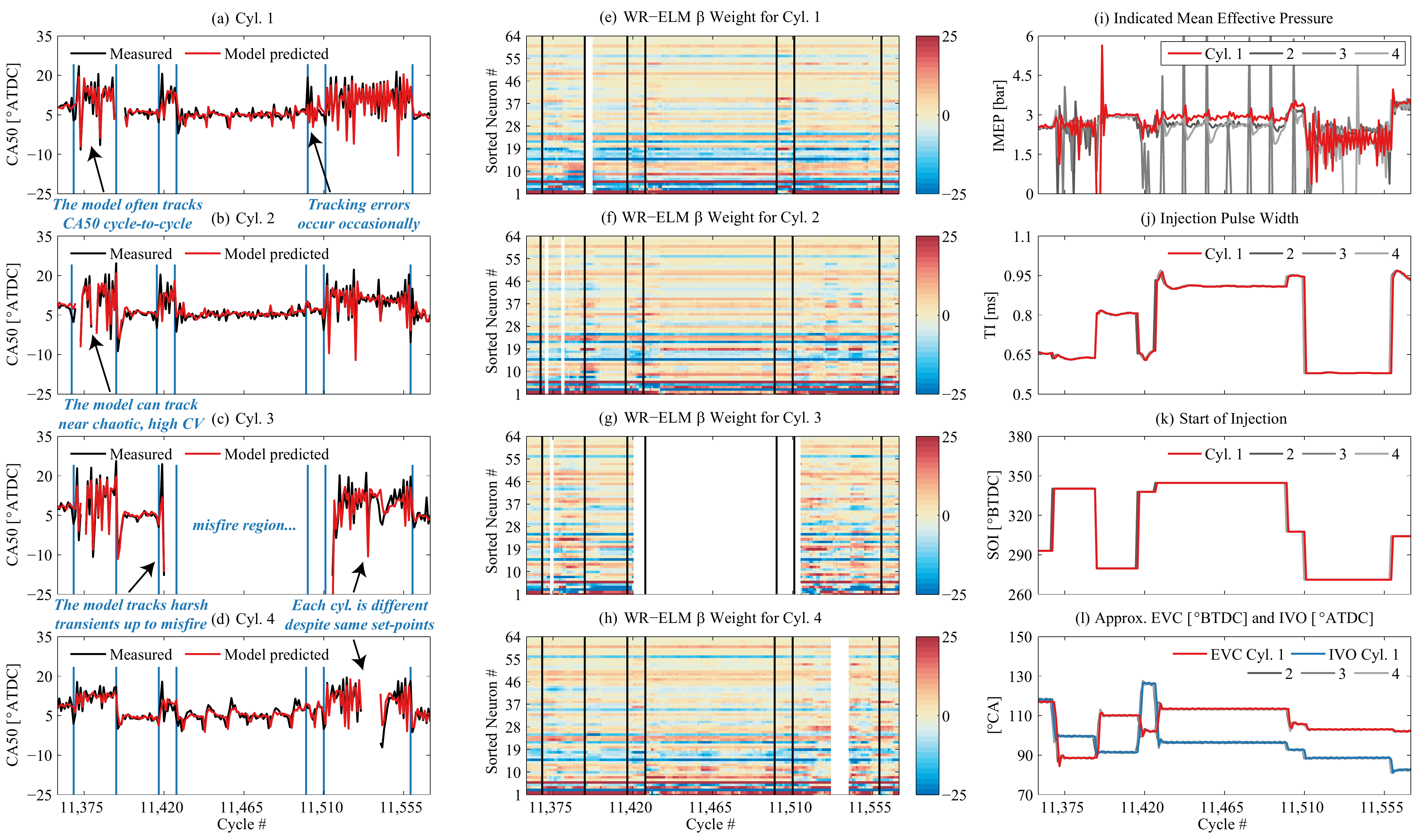}{1}{1.0}{0in 0in 0in 0in}{The WR-ELM CA50 model of Eq.~\ref{theModel} can track CA50 through transients every 0.5 - 10 seconds,  operating points with near chaotic high CV, and at steady-state during a particularly harsh region of the 25,323 cycle dataset that includes misfires. The colormaps (linearly scaled) provide qualitative insight into the level of cycle-to-cycle adaptation and into cylinder-to-cylinder model differences.}{betas}

\section{Summary and Conclusion}
This work presents a new online adaptation algorithm named Weighted Ring - Extreme Learning Machine.  The approach uses a weighted ring buffer data structure of recent measurements to recursively update an offline trained Extreme Learning Machine solution.  It is shown that WR-ELM can be used to approximate the combustion mapping function developed in~\cite{icefPaper} and provide reasonably accurate, causal predictions of near chaotic combustion behavior.  In the combustion application only in-cylinder pressure and crank encoder sensors are needed for predictions, and these predictions can be computed as early as 358 {\degrees}BTDC firing.  The algorithm is fast, and has been implemented in real-time on the low-cost Raspberry Pi\argh~platform (a two-minute video demonstrating this is available at~\cite{videoOfMyAlgorithm}).

Future work will explore optimal selection of weight(s) and try to better understand the situations that lead to the occasional model tracking errors.  Finally, the broader objective of this new modeling approach is to enable a new class of cycle-to-cycle model predictive control strategies that could potentially bring HCCI's low engine-out \ce{NO_x} and reduced \ce{CO2} emissions (higher fuel efficiency) to production gasoline engines.

\section*{Acknowledgments}
This material is based upon work supported by the Department of Energy [National Energy Technology Laboratory] under Award Number(s) DE-EE0003533. This work is performed as a part of the ACCESS project consortium (Robert Bosch LLC, AVL Inc., Emitec Inc., Stanford University, University of Michigan) under the direction of PI Hakan Yilmaz and Co-PI Oliver Miersch-Wiemers, Robert Bosch LLC.

The authors thank Vijay Janakiraman for providing the raw data analyzed in this paper.   The authors also thank Jeff Sterniak for both his test cell and project support.  A. Vaughan thanks Nisar Ahmed for many helpful discussions, and his advisors S. V. Bohac and Claus Borgnakke for the freedom to explore the interesting topic of near chaotic combustion.

\section*{Conflict of Interest}
The University of Michigan has filed a provisional patent on the work described in this publication.  A. Vaughan is named as the inventor, which includes royalty rights.  S. V. Bohac declares no conflict of interest.

\bibliographystyle{elsarticle-num-names}
{\raggedright {\footnotesize\bibliography{newPaper}}

{\setlength{\footnotemargin}{0pt}\let\thefootnote\relax\footnotetext{\scriptsize DISCLAIMER:  This report was prepared as an account of work sponsored by an agency of the United States Government. Neither the United States Government nor any agency thereof, nor any of their employees, makes any warranty, express or implied, or assumes any legal liability or responsibility for the accuracy, completeness, or usefulness of any information, apparatus, product, or process disclosed, or represents that its use would not infringe privately owned rights. Reference herein to any specific commercial product, process, or service by trade name, trademark, manufacturer, or otherwise does not necessarily constitute or imply its endorsement, recommendation, or favoring by the United States Government or any agency thereof. The views and opinions of authors expressed herein do not necessarily state or reflect those of the United States Government or any agency thereof.}
}

\end{document}